# Adaptability of Improved NEAT in Variable Environments

Destiny Bailey

Acadia University

February 19, 2021

## Abstract

A large challenge in Artificial Intelligence (AI) is training control agents that can properly adapt to variable environments. Environments in which the conditions change can cause issues for agents trying to operate in them. Building algorithms that can train agents to operate in these environments and properly deal with the changing conditions is therefore important. NeuroEvolution of Augmenting Topologies (NEAT) was a novel Genetic Algorithm (GA) when it was created, but has fallen aside with newer GAs outperforming it. This paper furthers the research on this subject by implementing various versions of improved NEAT in a variable environment to determine if NEAT can perform well in these environments. The improvements included, in every combination, are: recurrent connections, automatic feature selection, and increasing population size. The recurrent connections improvement performed extremely well. The automatic feature selection improvement was found to be detrimental to performance, and the increasing population size improvement lowered performance a small amount, but decreased computation requirements noticeably.

## Introduction

In artificial intelligence, Genetic Algorithms (GAs) are a way of training Neural Networks (NNs) to solve problems. GAs do this by creating successive generations of populations consisting of NNs that have a higher fitness score over time. These improvements occur by mutating the weights of connections between neurons in the NNs, and choosing the best of each generation to produce the next with. This paper will not further explain GAs, so here are some invaluable sources that explain GAs further: (Mitchell, Melanie (1996). An Introduction to Genetic Algorithms. Cambridge, MA: MIT Press., Goldberg, David (1989). Genetic Algorithms in Search, Optimization and Machine Learning. Reading, MA: Addison-Wesley Professional., Holland, John (1992). Adaptation in Natural and Artificial Systems. Cambridge, MA: MIT Press.). NeuroEvolution of Augmenting Topologies (NEAT) is a GA, however, it mutates the topology of the network along with the weights of the connections, changing where connections and neurons exist [17].

NNs have applications in the real world solving problems too complex for traditional programming to handle well. Many problems in the real world involve environments that rapidly change and require adaptability from the NN. GAs are a prominent form of training NNs to solve these problems. NEAT was made as an improved form of GA and is still seen as having potential. Newer GAs have come after NEAT that can be applied to the same problems [3, 4].

NEAT was created to perform better than other GAs on complex problems that they have trouble with. NEAT was originally tested on the double-pole balancing problem and performed better than all other GAs it was tested against. NEAT has also been successfully applied to: geological predictions, car racing, automobile crash warning systems, video game controllers, and many more problems. Many of these problems have rapidly varying environments that require adaptability from the NN in order for it to perform well [1, 11, 13, 17, 19].

In recent tests with problems that require adaptability, NEAT has been consistently outperformed. In these tests, NEAT has been tested against newer GAs, which always perform better in these adaptability problems. This is concerning for the relevance of NEAT, as if there are better GAs, no one will use NEAT as an effective GA [9, 10, 12].

In the tests of NEAT with adaptability problems, the type of NEAT used is always the original version. NEAT was one of the best GAs when it was created, and it is still seen as having potential because of its novelty. NEAT, in its original version, does not seem able to compete with newer GAs.

Over time since its creation, NEAT has undergone various improvements from various sources. These improvements exist, yet are not used other than in the experiment for which they were designed. These improved versions of NEAT have the potential to perform better than original NEAT in problems that require adaptability [12, 14, 19, 20, 21].

**Methodology**

To test the adaptability of NEAT an experiment will be conducted to implement multiple improved NEAT algorithms to train NNs to perform in the dangerous foraging domain. The training will involve using consistent parameters for implementing all versions of improved NEAT, to ensure standardized results. The training will be reviewed for champion fitness and other important variables for the chosen version of improved NEAT. That version of improved NEAT will be tested five times to allow for a better average to be taken. All the training data for that version of improved NEAT will be analysed and conclusions will be made from the data. Other versions of improved NEAT will then be used to determine the maximum efficacy of each.

The versions of improved NEAT being tested have two goals: increasing performance, and decreasing computational requirements. Decreasing computational requirements involves lowering the computing power required for the algorithm to run, this is not the time it requires to run as a more powerful computer could run an algorithm faster than another less powerful computer. Increasing performance involves a faster improvement of the fitness score without increasing computational requirements.

The three versions of improved NEAT being tested are: recurrent connections, automatic feature selection, and increasing population size. Automatic feature selection starts every neural network with only one connected input. Theoretically, this allows for the training algorithm to connect only the required inputs, and leave the unnecessary inputs unconnected. This should increase performance, but not computational requirements. The vast majority of computational power goes to running the simulations, so removing small parts of the network topology should not change the computation requirements. The increase in performance comes from the algorithm not starting with unnecessary inputs, allowing it to ignore them if it forms a connection to them. The unimproved NEAT algorithm has the ability to drop any connections, including inputs, which may make automatic feature selection ineffective in improving performance. Increasing population size starts the population at size 200, increasing it to the full size 500 over the generations. There should be minor performance impact as even with a smaller starting population, finding definitive species should still happen, and fine-tuning solutions with the larger population at the end of the generations still occurs. There should be a large reduction in computational requirements as many fewer simulations are being run at the beginning. Recurrent connections allow neurons to communicate temporally, increasing performance by a very large amount.

The dangerous foraging domain presents a problem that must be overcome with adaptability. The dangerous foraging domain was first presented in *Evolving Adaptive Neural Networks with and without Adaptive Synapses* by Kenneth O. Stanley, Bobby D. Bryant, and Risto Miikkulainen [12]. The dangerous foraging domain involves the trained NN controlling a simulated robot in a field with eight items of food dispersed randomly through the field, there being two possible types of food, A and B. The NN must control the robot to eat both kinds of food until it discovers which kind of food is poisonous, it must then stop foraging that type of food. To test a NN it will go through eight simulations: two simulations with type A being edible, two simulations with type B being edible, two simulations with type A being poisonous, and two simulations with type B being poisonous. In this case, the testing of adaptability is done by requiring the NN to change its policy in each simulation. The NN must be able to forage for both kinds of food in every simulation, until it finds out if that kind of food is poisonous, which it will then avoid for the rest of that simulation. The ability for a NN to change policy so rapidly is a good display of adaptability, and the dangerous foraging domain presents a great variable environment; so it was chosen for this experiment.

The data gathered from the experiment will be analysed for the total number of generations until consistent champion success is achieved, as well as other important aspects. The total number of generations until consistent champion success is a very good indication that the GA has trained the NN to a solution to the problem, as it represents that the best NN trained is able to consistently succeed at finding a solution to the problem. The computational requirements of the training algorithm is also an important consideration. If an algorithm has less computational requirements per generation, more generations can be run to find a solution. This is an important consideration as my training had a set generation limit at which it stopped, so more efficient algorithms could have run longer to match the total computation of the less efficient algorithms. Other important aspects to analyse would be the diversity of species and network topology; however, these are hard to get a lot of useful information from as the way the algorithm operates is mostly a black box and these reflect that.

Fitness is calculated by tracking the number of edible foods eaten and the number of poisonous foods eaten, and comparing them. The fitness is calculated by taking the sum of all edible food eaten, with poisonous food eaten taking away from that sum. As four of the eight simulations are poisonous, the maximum poisonous food eaten is thirty-two. To ensure positive fitness, the fitness function $f$ is defined,

$$f = 32 + e - p,$$

where $e$ is the number of edible items consumed and $p$ is the number of poisonous items. This means that over all eight simulations the maximum fitness will be sixty-four, but since four of the trials are poisonous and the perfect solution would require eating one from each, the actual maximum fitness function is sixty.

The simulation is made up of timesteps. Each timestep, the distance to the closest food of each type is obtained from sensors on: the front left, front centre, front right, back left, and back right of the robot. If a food type is not present on a simulation, which will occur each simulation, all five sensors for that type return zero. There is one sensor for pleasure and one for pain. If the robot eats a poisonous food, the pain sensor is activated for twenty time steps. If an edible food is eaten, the pleasure sensor is activated for twenty time steps. There is also a biased input. All thirteen inputs are gathered each timestep and put into the NN currently being tested. The NN has three outputs: left rotation, right rotation, and forward movement. Each timestep, those movements are applied and food eaten is checked. After moving, if the robot is overlapping any foods, they become eaten. Eaten foods stop appearing on sensors, can only be eaten once, and affect the fitness score. Each simulation, the robot starts in the middle of the field with the eight food items placed randomly in the field.

Due to the random nature of placement and food possibly overlapping or starting overlapping the robot, it can be easier to get more food faster than usual. The randomness makes it possible to achieve a better score if it is an edible simulation, or a lower score if it is a poisonous simulation. This randomness of placement is an issue that could be fixed, not allowing starting overlaps. Fixing the issue would improve the accuracy of the dangerous foraging domain to evaluate NNs, but the issue is very minor and was deemed acceptable in this paper.

Each simulation starts with the robot placed in the middle of the field and the food is placed randomly. Each timestep, inputs are put through the NN being tested, and the outputs of the NN are applied to the simulation to move the robot. The robot moves and food eaten is checked. The simulation lasts for a set number of timesteps, in this paper there are 750 timesteps every simulation, then fitness is calculated.

The data gathered from the experiment will allow the point of consistent champion success to be determined. Consistent champion success is determined by observing the champion of each generation. The champion is the member of the population that achieved the highest fitness that generation. When the champion has achieved the highest possible fitness for multiple generations in a row, consistent champion success has been achieved. This is a very good identifier that the GA has trained the population to the point where the best NN is always solving the problem, which means training has been successful. The time to train will be monitored for total time taken for each consistent champion success achieved, but since the machine used for the training will be undergoing other uses during training, the time taken for training could possibly vary due to this and is therefore not reliable.

## Experiments

Eight total tests were run to test the efficacy of improved NEAT. Each test consisted of five sets of simulations to allow for a better average of the performance of a specific algorithm to be obtained, all results in this paper referencing a specific algorithm tested will refer to the average for that algorithm over those five simulations. Each set of simulations was 500 generations with a population size of 500. Each generation, every NN was placed in the dangerous foraging domain controlling the robot. The dangerous foraging domain lasted 750 timesteps each simulation.

All simulations were run on a machine under other loads, therefore the exact time required by each simulation is not usable data as the machine each simulation was run on was under other varying loads at those times. All of the simulation sets were the various combinations of the three versions of improved NEAT being tested, including a set with none of them.

The parameter settings for all GAs tested were those that the NEAT-Python library had set for the single pole balancing example problem. Determining parameter settings through trial and error for this specific implementation of NEAT was deemed beneficial, but due to the challenges involved, unnecessary for the purposes of this paper. Due to issues with the stagnation of top-performing species and having them removed after twenty generations, the top three performing species were never removed due to stagnation. All simulations ran for 500 generations, regardless if they reached the fitness threshold, to allow for the consistent champion success to be better determined.

**Results**

The first test was of the original NEAT algorithm. The original NEAT algorithm was not meant to perform well in the dangerous foraging domain. The dangerous foraging domain was designed for a specifically adaptive NEAT algorithm to perform well, so the original NEAT algorithm was expected to perform poorly. The original NEAT algorithm performed as expected, with champion fitness usually under 50 (figure 1). The original NEAT algorithm was run as a control, to compare all of the improved NEAT algorithms against. Having a control was deemed useful for comparison, and was not difficult to accomplish. This data on the performance of the original NEAT algorithm will prove useful for determining if the improved NEAT algorithms were successful in improving performance.

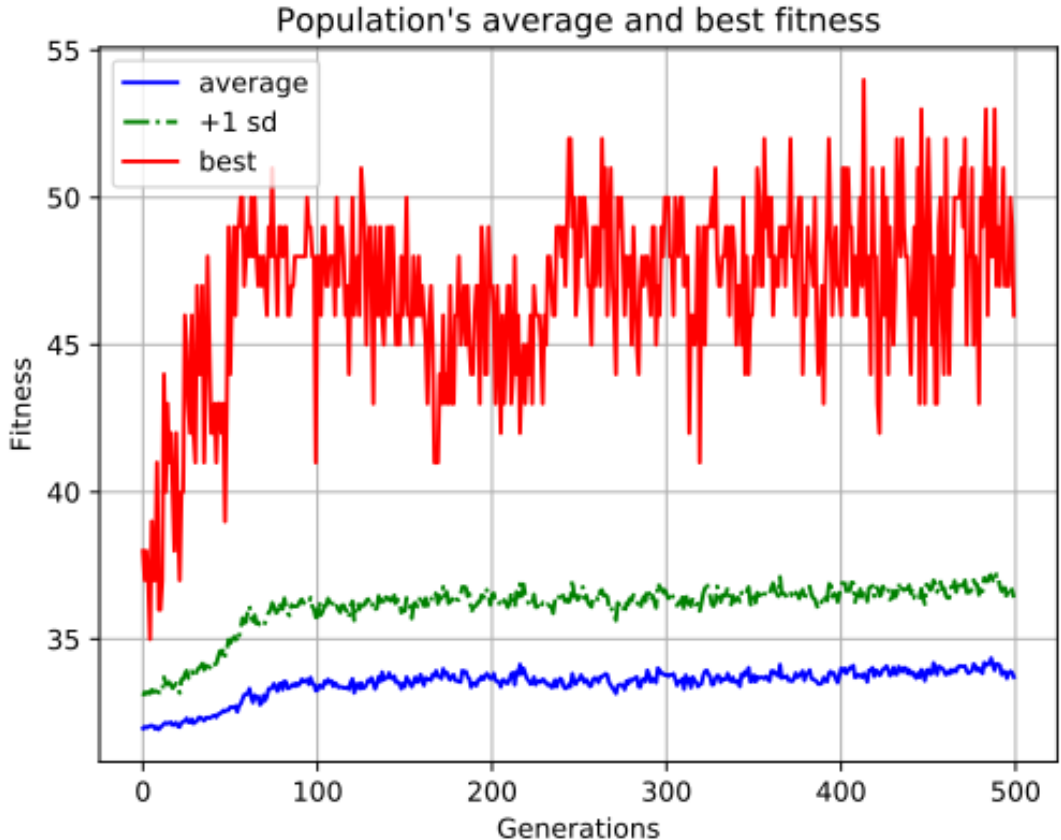

Figure 1: **Example original NEAT fitness graph.** The original NEAT champion fitness scores were very erratic.

The fitness graphs are fitness plotted over generations. There are three lines representing fitness. The red line is the champion fitness for that generation. The blue line is the average fitness for that generation. The green line is the average fitness plus one standard deviation for that generation.

The second test was of the increasing population size improvement alone. There was not much visible performance difference between the increasing population size improvement and the original NEAT algorithms. They were both very erratic and failed to find a solution. The fitness score was usually under 50 but changed unpredictably (figure 2). These results were expected. The very similar performance between the increasing population size improvement and the original NEAT algorithms shows that, in this situation, there is no noticeable performance difference between the two. This was promising for the increasing population size improvement's efficacy at decreasing computational requirements while not decreasing performance. There was no noticeable decrease in computational requirements compared to the original NEAT algorithm, but it was very difficult to monitor the computational requirements on the machine the simulations were being run on, so there is no data for the computational requirements of the increasing population size improvement alone.

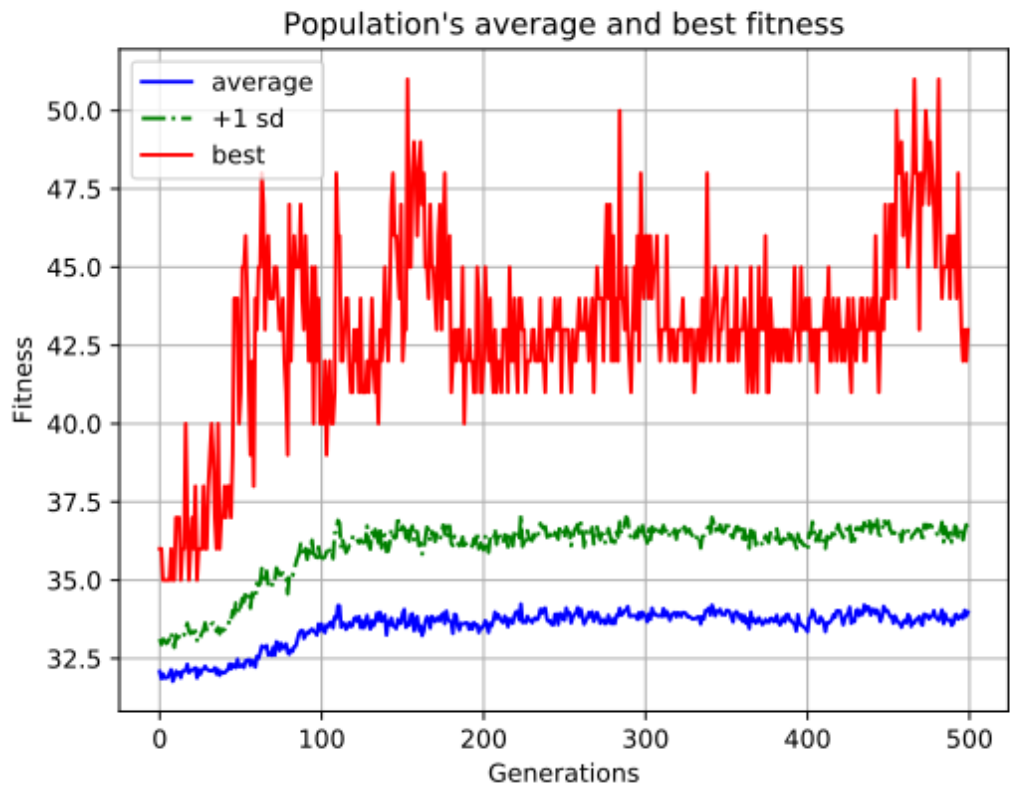

Figure 2: **Example increasing population size fitness graph.** Similar to most of the original NEAT simulations, the champion fitness was very erratic. While the fitness score rarely went above 50, when it did, it would often drop to a lower value.

The third test was of the automatic feature selection improvement alone. The results were not as expected. The automatic feature selection improvement is supposed to start with only one connected input, and attach other inputs only as needed. This behaviour of only connecting necessary inputs, in theory, improves performance by connecting only necessary inputs and never having detrimental inputs connected. The performance of automatic feature selection was similar to that of the original NEAT algorithm. The fitness score stayed below 50 most of the time and no solution was found (figure 3). This could be because the original NEAT algorithm starts with only some inputs connected, not fully connected, which could be doing what automatic feature selection is doing, but better. The other simulations with automatic feature selection will show similar results.

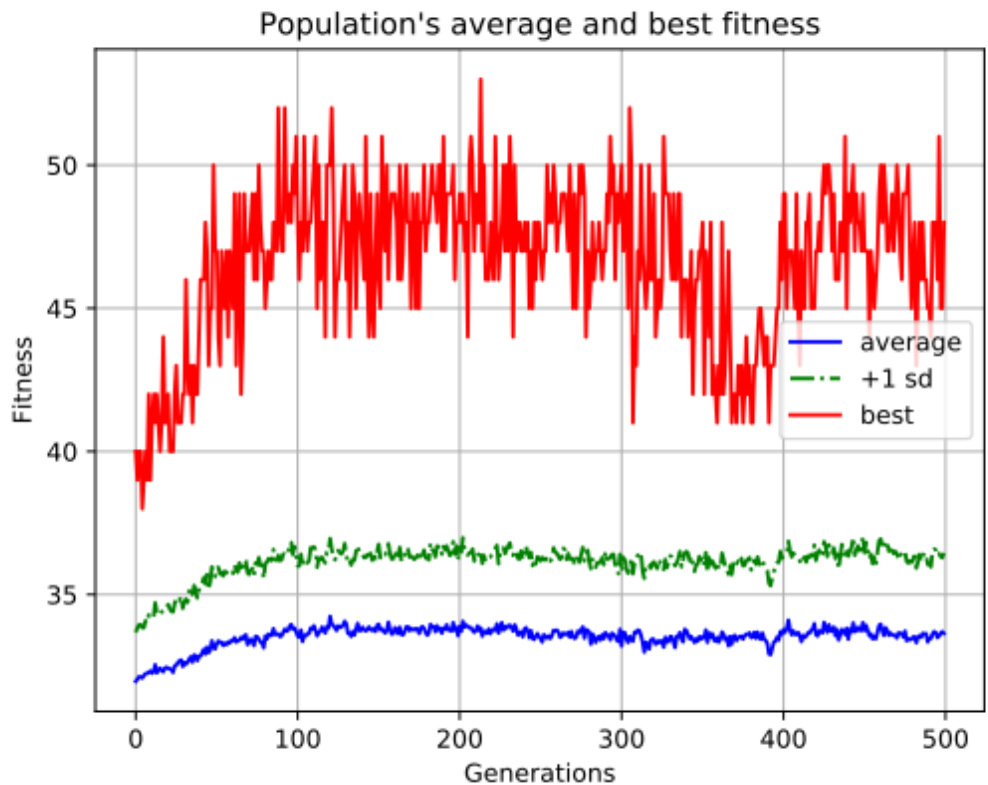

Figure 3: **Example automatic feature selection fitness graph.** The fitness score remained lower than 50 most of the time, and no solutions were ever found. There was also erratic behaviour with the fitness score dropping and rising commonly by large amounts.

The fourth test was of the recurrent connections improvement alone. The results of the recurrent connections improvement were very surprising. All simulations with the recurrent connections improvement alone found a solution around 175 generations (figure 4). These results were surprising because the performance increase was so extreme. The recurrent connections improvement was supposed to increase performance, but going from original NEAT having a fitness score usually below 50 and finding no solutions, to having a fitness score usually above 60 and finding a solution around 175 generations is an extreme difference. The recurrent connections improvement alone performed the best of all simulation sets performed.

The results were also surprising for another reason. In *Evolving Adaptive Neural Networks with and without Adaptive Synapses*, the fixed-topology algorithm that performed extremely well compared to the NEAT algorithm with adaptive synapses did worse than the recurrent connections improvement alone tested in this paper. The NEAT algorithm with adaptive synapses, on its best run, found a solution in 350 generations. The fixed-topology algorithm found a solution on its best run in 250 generations, consistently finding a solution by 350 generations [12]. For the recurrent connections improvement alone, tested in this paper, to perform much better than the fixed-topology algorithm, which performed surprisingly well in that paper, was extremely surprising.

The recurrent connections improvement alone took noticeably longer to run compared to the original NEAT algorithm. The longer run time means that the recurrent connections improvement had much larger computational requirements compared to the original NEAT algorithm. Due to the machine these simulations were running on being under other loads, it is not possible to draw any specific data from the time taken to run. The only usable data relating to the time taken to run is when there is an extreme difference in time taken. The larger computational requirements of the recurrent connections improvement would suggest a better fitness score, as more computation power is being used to run it.

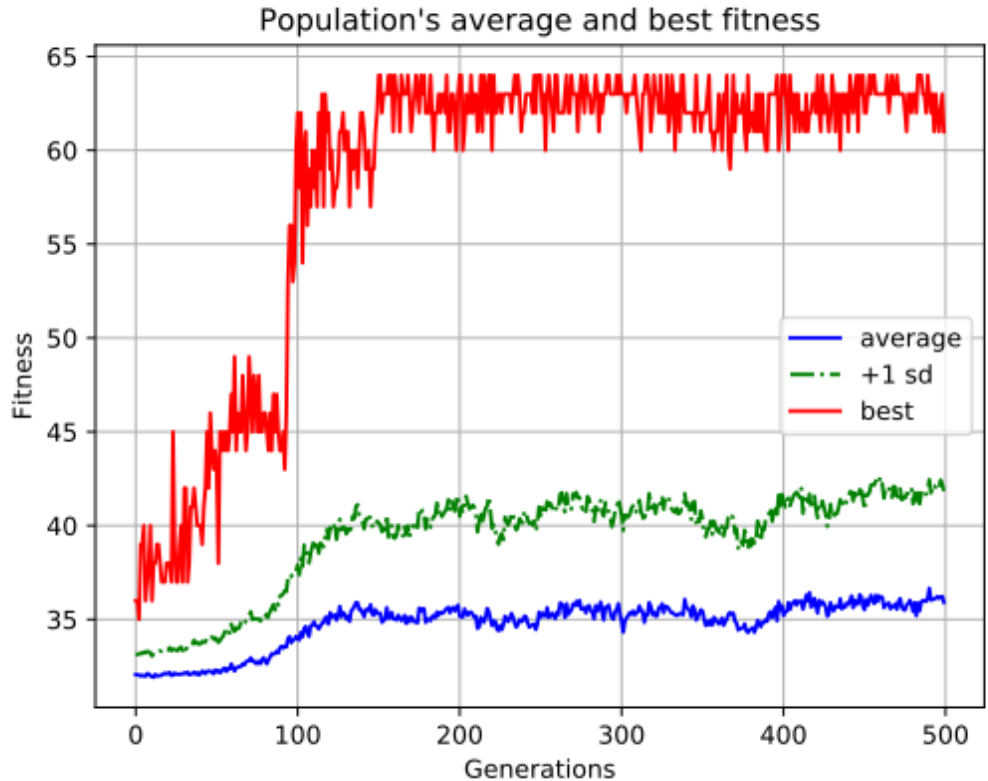

Figure 4: **Example recurrent connections fitness graph.** This example was the best, finding a solution just after 100 generations, a very good performance.

The fifth test was of the automatic feature selection and increasing population size improvements together. These simulations performed the worst compared to all other simulations performed. The fitness score was under 50 most of the time, and no solutions were found (figure 5). The difference in performance between the automatic feature selection and increasing population size improvements together and the original NEAT algorithm is very small. This slight decrease in performance could be amplified when applied with the recurrent connections improvement, or it could remain the same or have little impact on performance. The simulation set with all improvements shows a severe impact on performance. As with the increasing population size alone, there was no data on computational requirements for this simulation set.

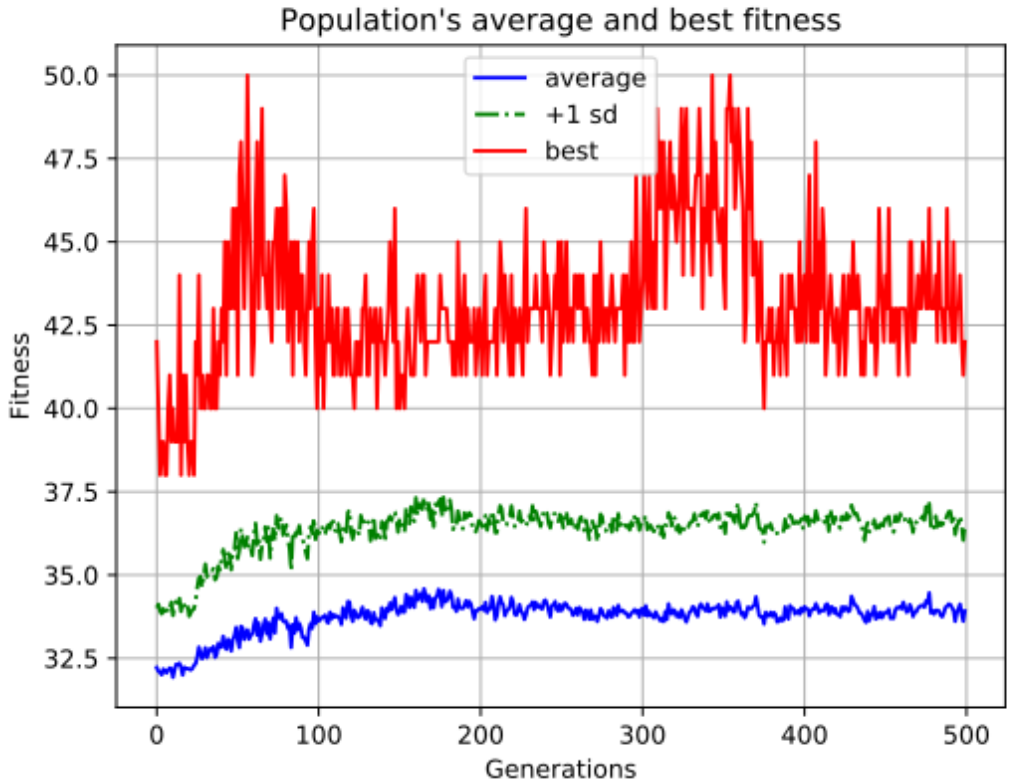

Figure 5: **Example automatic feature selection and increasing population size improvements fitness graph.** The fitness score in all simulations of the automatic feature selection and increasing population size improvements together was very erratic. The fitness score was mostly below 50, rarely going above.

The sixth test was of the recurrent connections and increasing population size improvements together. Of the five simulations, four found solutions around 200 generations, and one failed to find a solution (figure 6). This is a good example of why multiple simulations were run to get a better average of the performance of the algorithms, as most were successful, with seemingly an outlier. The performance of the successful simulations is very promising for the efficacy for the increasing population size improvement, as increasing from 175 generations to 200 generations to find consistent champion success is not a large loss in performance. The one simulation that failed to find a solution seems to not be a good sign for the efficacy of the increasing population size improvement, but it could be an outlier case, so before any conclusions are made about this being an issue more simulations need to be run, which was not done for this paper. The time taken for these simulations was noticeably shorter than that of the recurrent connections improvement alone, which is good for the efficacy of the increasing population size improvement.

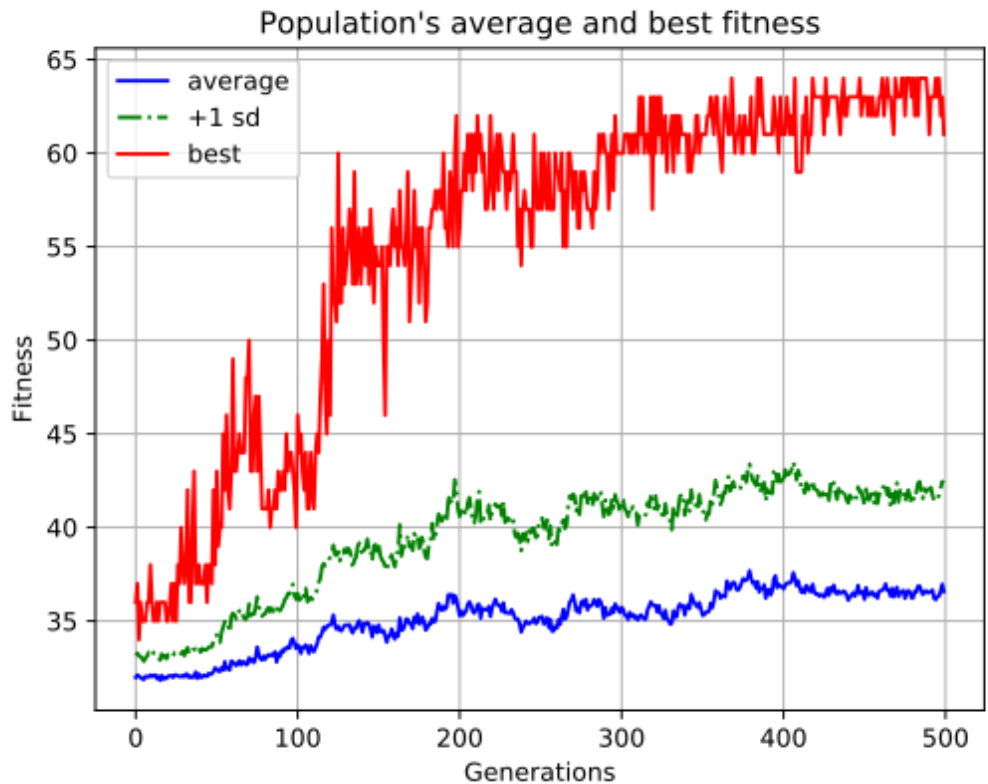

Figure 6: **Example recurrent connections and increasing population size fitness graph.** The performance of the successful simulations was very good, finding solutions by around 200 generations. Finding a solution by 200 generations is only 25 generations slower than that of the recurrent connections improvement alone!

The seventh test was of the recurrent connections and automatic feature selection improvements together. The results were surprisingly bad. Of the five simulations, one found a solution by around 200 generations, another one found a solution by around 300 generations, and one found a solution by around 375 generations, the last two did not find solutions (figure 7). Compared to the drop in performance of the automatic feature selection improvement alone, which was not noticeable under inspection, this drop in performance is very extreme. The recurrent connections improvement seemed to have a very hard time finding a solution with the automatic feature selection improvement active. The solutions found were no better than those found by the recurrent connections improvement alone. The bad performance is not good for the efficacy of the automatic feature selection improvement.

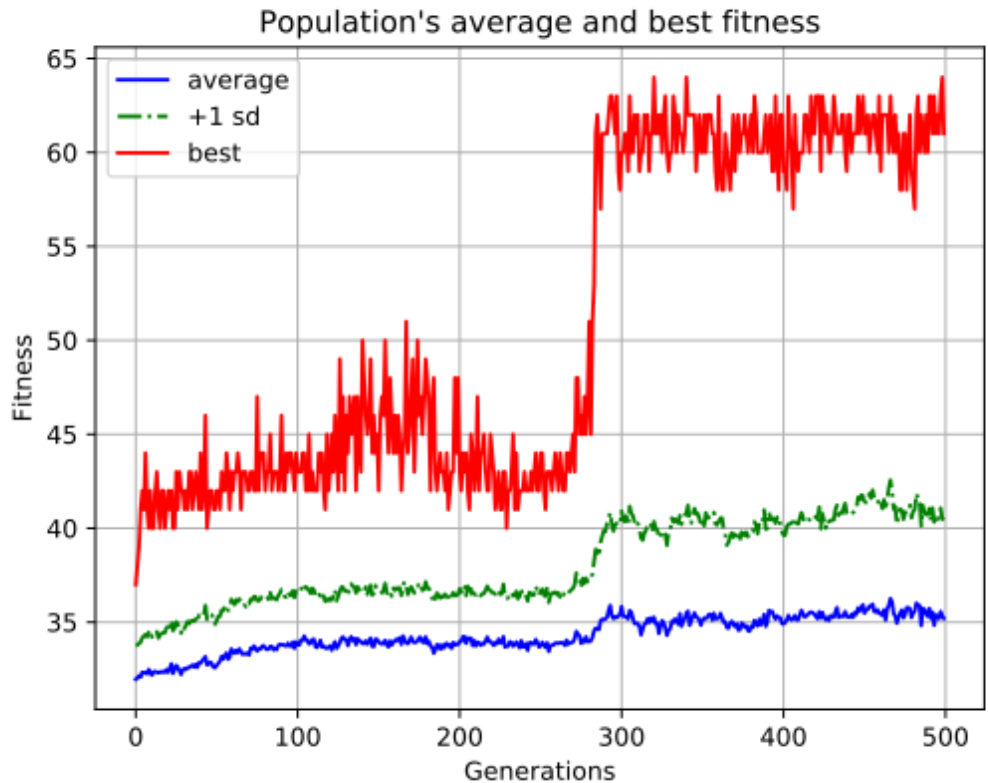

Figure 7: **Example recurrent connections and automatic feature selection improvements fitness graph.** Due to the erratic nature of these simulations, there is no good single example. This example is effective at showcasing the erratic nature of this set of improvements by having a very sudden spike in the fitness score just before 300 generations, which found a solution.

The eighth test was of the recurrent connections, automatic feature selection, and increasing population size improvements together. The results were as expected, the performance was quite poor. Of the five simulations, two found solutions by around 250 generations, another two found solutions by around 350 generations, and the last one failed to find a solution. The results were expected to be worse, as the automatic feature selection and recurrent connections improvements together performed worse than these simulations when these simulations also have the increasing population size improvement, which should have decreased performance further. This is a good example of how erratic the automatic feature selection improvement is in this environment, as even with another improvement that should have decreased performance more, it happened to increase. The recurrent connections, automatic feature selection, and increasing population size improvements together performed poorly, which is not good for their efficacy.

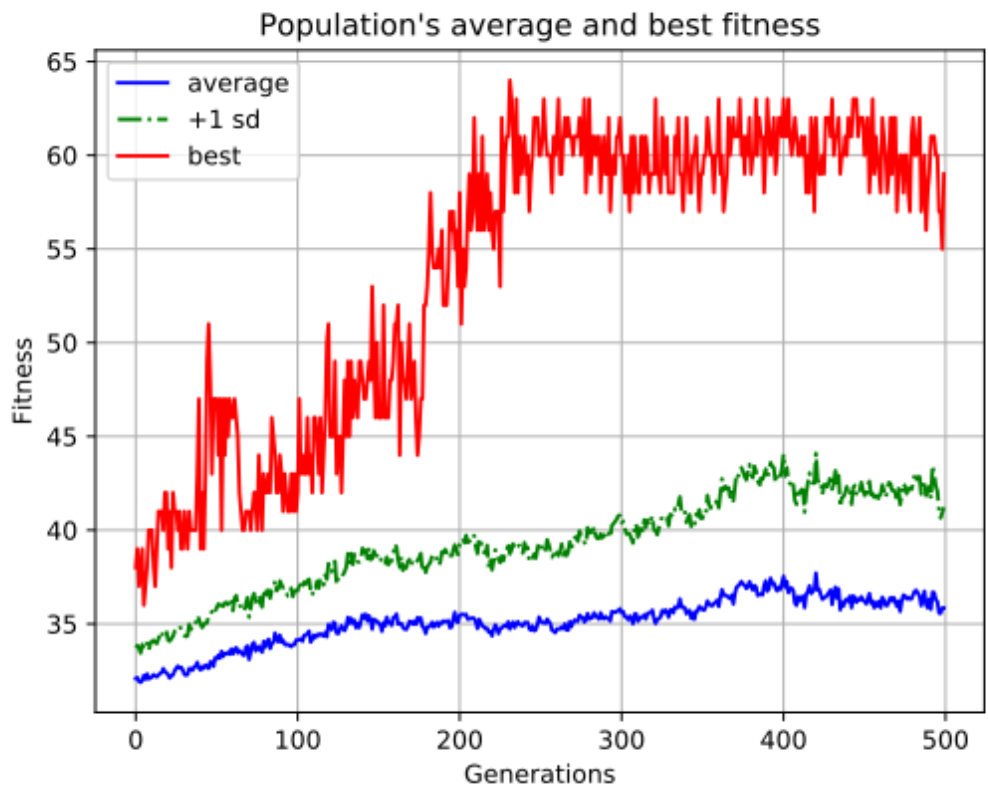

Figure 8: **Example recurrent connections, automatic feature selection, and increasing population size fitness graph.** In this example, a solution is found just before 250 generations. This was the best simulation. Even though these simulations performed better on average compared to the automatic feature selection and recurrent connections improvements together, the automatic feature selection improvement has shown to be very erratic in this environment, making these better scores unlikely to be a good representation of its performance.

**Conclusion**

The original NEAT algorithm along with the automatic feature selection, increasing population size, and recurrent connections improvements, were all tested in the dangerous foraging domain. The task required adaptive behaviour by the NNs during simulations. The original NEAT algorithm found no solutions. The increasing population size and automatic feature selection improvements alone found no solutions. The recurrent connections improvement alone performed exceedingly well compared to what was expected based on the original NEAT algorithm and the results presented in *Evolving Adaptive Neural Networks with and without Adaptive Synapses*. The automatic feature selection improvement is not a useful improvement in this environment, as partial starting connections find solutions faster. The increasing population size improvement was faster computationally, but took more generations to find a solution. The efficacy of both the recurrent connections improvement alone and the recurrent connections and increasing population size improvements together are very promising.

The solutions found by the fixed-topology algorithm in *Evolving Adaptive Neural Networks with and without Adaptive Synapses* were concluded to be from recurrency alone. The recurrent connections were determined to be effective at finding easy solutions to the dangerous foraging domain [12]. The solutions found by the NEAT algorithm with the recurrent connections improvement were much better than the solutions of the fixed-topology algorithm presented in *Evolving Adaptive Neural Networks with and without Adaptive Synapses*, finding solutions around 100 generations faster and having a higher average champion fitness score. This could show that NEAT is an improvement over the fixed-topology algorithm, but this difference in performance could also be due to the difference in time between the date of these studies, almost 20 years, and the level of technological development could be completely different.

**Future Work**

Running the simulation sets on a machine not under other loads to determine the difference in run times between the recurrent connections improvement alone and the recurrent connections and increasing population size improvements together by determining which algorithm found a solution first would provide invaluable data on the efficacy of the increasing population size improvement. Comparing more versions of improved NEAT and other GAs would provide more information on the efficacy of NEAT.

## Acknowledgements

This paper is an abridged version of an unpublished paper of the same name written by myself in May 2020. My AP Research 12 teacher, Floyd Priddle, was very helpful with his writing instruction. My professor, Dr. Andrew Mcintyre, gave minor editorial suggestions. The NEAT-Python library was used for all experiments in this paper and was invaluable.

## Key Terms

- Artificial intelligence, machine learning
- Genetic algorithm, evolutionary algorithm
- Neuroevolution, neural network
- Adaptability, variable environments
- Evolution strategies, evolutionary programming
- Network topology, network structure, connection weights,
- Fitness, evaluation
- Speciation, preserve innovation
- Increasing population size
- Automatic feature selection
- Recurrent connections, recurrent neural network